\documentclass[10pt,twocolumn,letterpaper]{article}

\usepackage{cvpr}
\usepackage{times}
\usepackage{epsfig}
\usepackage{graphicx}
\usepackage{amsmath}
\usepackage{amssymb}
\usepackage{amsfonts}            
\usepackage{mathrsfs} 
\usepackage{booktabs}
\usepackage{multirow}          
\usepackage{color}
\usepackage{subfigure}
\usepackage{cite}
\usepackage{verbatimbox}
\usepackage[breaklinks=true,bookmarks=false,colorlinks]{hyperref}

\cvprfinalcopy 

\def\cvprPaperID{8350} 

\begin{document}

\title{Distribution-induced Bidirectional Generative Adversarial Network for Graph Representation Learning}
\author{Shuai Zheng\textsuperscript{1,2}, Zhenfeng Zhu\textsuperscript{1,2,\thanks{Corresponding author.}}, Xingxing Zhang\textsuperscript{1,2}, Zhizhe Liu\textsuperscript{1,2}, Jian Cheng\textsuperscript{3,4}, Yao Zhao\textsuperscript{1,2} \\
\textsuperscript{1}Institute of Information Science, Beijing Jiaotong University, Beijing, China\\
\textsuperscript{2}Beijing Key Laboratory of Advanced Information Science and Network Technology, Beijing, China\\
\textsuperscript{3}NLPR, Institute of Automation, Chinese Academy of Sciences, Beijing, China\\
\textsuperscript{4}University of Chinese Academy of Sciences, Beijing, China\\
{\tt\small \textsuperscript{1,2}\{zs1997,zhfzhu,zhangxing,yzhao\}@bjtu.edu.cn,  \textsuperscript{3,4}jcheng@nlpr.ia.ac.cn }
}

\maketitle
\thispagestyle{empty}
\pagestyle{empty}  
\begin{abstract}  
	Graph representation learning aims to encode all nodes of a graph into low-dimensional vectors that will serve as input of many compute vision tasks.
	However, most existing algorithms ignore the existence of inherent data distribution and even noises. This may significantly increase the phenomenon of over-fitting and deteriorate the testing accuracy.
	In this paper, we propose a \underline{D}istribution-induced \underline{B}idirectional \underline{G}enerative \underline{A}dversarial \underline{N}etwork (named \textbf{DBGAN}) for  graph representation learning. 
	Instead of the widely used normal distribution assumption, the prior distribution of latent representation in our DBGAN is estimated in a structure-aware way, which implicitly bridges the graph and feature spaces by prototype learning. Thus discriminative and robust representations are generated for all nodes.
	Furthermore, to improve their generalization ability while preserving representation ability, the sample-level and distribution-level consistency is well balanced via a bidirectional adversarial learning framework.
	An extensive group of experiments are then carefully designed and presented, demonstrating that our DBGAN obtains remarkably more favorable trade-off between representation and robustness, and meanwhile is dimension-efficient, over currently available alternatives in various tasks. The source code is released in \url{https://github.com/SsGood/DBGAN}.
\end{abstract}

\section{Introduction}
A graph is a collection of nodes and edges that can be used to model relationships and processes between data in a variety of scenarios, such as biomedical networks, citation networks, and social networks. Therefore, graph analysis is a necessary step to explore the internal information of these networks. 
However, due to the complex topology and high data dimension of graph data, most of the current machine learning methods for simple sequences or grids design are not suitable for graph data analysis. 
As a general approach to these problems, Graph representation learning aims to represent sparse raw features of graph nodes as compact low-dimensional vectors while preserving enough information for subsequent downstream tasks, such as link prediction\cite{GraphSGAN,graphgan}, clustering~\cite{mgae,spectral}, and recommendation~\cite{social, deepwalk}.
In recent years, a variety of graph representation learning methods have been proposed, which can be broadly summarized into two categories: proximity-based algorithms and deep learning-based algorithms. 

By applying matrix factorization, proximity-based algorithms, such as GraRep\cite{grarep}, HOPE\cite{HOPE}, M-NMF\cite{M-NMF} attempt to factorize the graph adjacency matrix to obtain the node representation. While for probabilistic models, such as DeepWalk\cite{deepwalk}, line\cite{LINE}, and node2vec\cite{node2vec}, they learn the node representation with local neighborhood connectivities through randomwalk and various order proximities. These methods are all focused on preserving the original neighborhood relationship in a low dimensional space. Recent studies have also shown that probabilistic models and matrix factorization-based algorithms are equivalent and can be implemented by a unified model\cite{qiu2018network}. 

Deep learning-based approaches are receiving increasing attention, most of which use the auto-encoder framework to capture the latent representation. SDNE\cite{SDNE} and DNGR\cite{DNGR} use deep auto-encoders to model the positive point-wise mutual information (PPMI) while preserving the structure of the graph. The GAE\cite{GAE} first merges the GCN\cite{GCN} as an encoder into the auto-encoder framework to seek the latent representation  by reconstructing the adjacency matrix. In addition, MGAE\cite{mgae}, GDN\cite{GDN}, and GALA\cite{GALA} attempt to preserve node feature in latent representation by building learnable decoders and encoders on a GAE basis. In fact, most of the above methods are to reconstruct either the adjacency matrix or the node feature, rather than the reconstruction on both together. However, for good low-dimensional latent representations, the topology of the graph and the node feature should be preserved at the same time.

It is worth noting that none of the above methods have explicitly exploited the latent distribution of the graphical data, and thus, the distribution consistency across domains(graph space and feature space) cannot be well preserved, which leads to poor generalization of the representation and sensitivity to noise. Due to the strong ability of the generative adversarial network(GAN)\cite{GAN} for distribution fitting, some works have introduced adversarial learning into the field of graph representation learning to improve the performance of the learned latent representation.  In GraphGAN\cite{graphgan} and ProGAN\cite{ProGAN}, the generated fake node pairs and node triplets compete with the real data to enhance the robustness of latent representation. These methods ignore the global structure and node feature, and fail to preserve the distributed consistency, resulting in the insufficiency in generalization ability. Besides, the normal distribution $\mathbb{N}(0,1)$ has been generally pre-assumed in AIDW\cite{AIDW} and ARGA\cite{ARGA} to guide the generation of latent representations. However, in most cases, it is not suitable to model the latent distribution of graph data by $\mathbb{N}(0,1)$, and an inaccurate prior distribution can cause the model to be over-smoothing or even misleading.

Motivated by the observations mentioned above, we propose a distribution-induced bidirectional GAN for unsupervised graph representation learning, named as DBGAN. To enhance the generalization ability of the representation, different from unidirectional mapping of data to representation in ARGA\cite{ARGA} and AIDW\cite{AIDW}, we not only apply adversarial learning to the encoder but also construct a generator for modeling the mapping of latent representation to graph data, establishing a bidirectional mapping between the two spaces, thus, the distribution consistency and sample consistency of the node representations are preserved in the latent space.
Furthermore, to preserve the structural consistency of graph data, we perform prior distribution estimation in latent space using the learned cross-domain prototypes. This will facilitate the robustness of node representations and alleviate the over-smoothing problem caused by normal distribution assumption like in ARGA\cite{ARGA}.
We evaluate the effectiveness of latent representations learned by GBGAN on both link prediction and node clustering tasks.
The contributions are highlighted in the following aspects:
\begin{itemize}
	\item We propose a \underline{D}istribution-induced \underline{B}idirectional \underline{G}enerative \underline{A}dversarial \underline{N}etwork (DBGAN), for graph representation learning with a dimension-efficient property. To the best of our knowledge, it is the first work to consider prior distribution estimation in adversarial learning.
	\item
	To improve generalization ability while preserving representation ability, the sample-level and distribution-level consistency are well balanced via bidirectional adversarial learning.
	\item 
	Unlike the widely used normal distribution assumption, we innovatively estimate structure-aware prior distribution of latent representation by bridging the graph and feature spaces with learned prototypes, thus generating robust and discriminative representations.
	\item
	Significant improvements over currently available alternatives demonstrate that our DBGAN creates a new baseline in the area of graph representation learning.
\end{itemize}
\section{GANs for Representation Learning}

\begin{figure}[t]	
	\centering
	\includegraphics[width=3in]{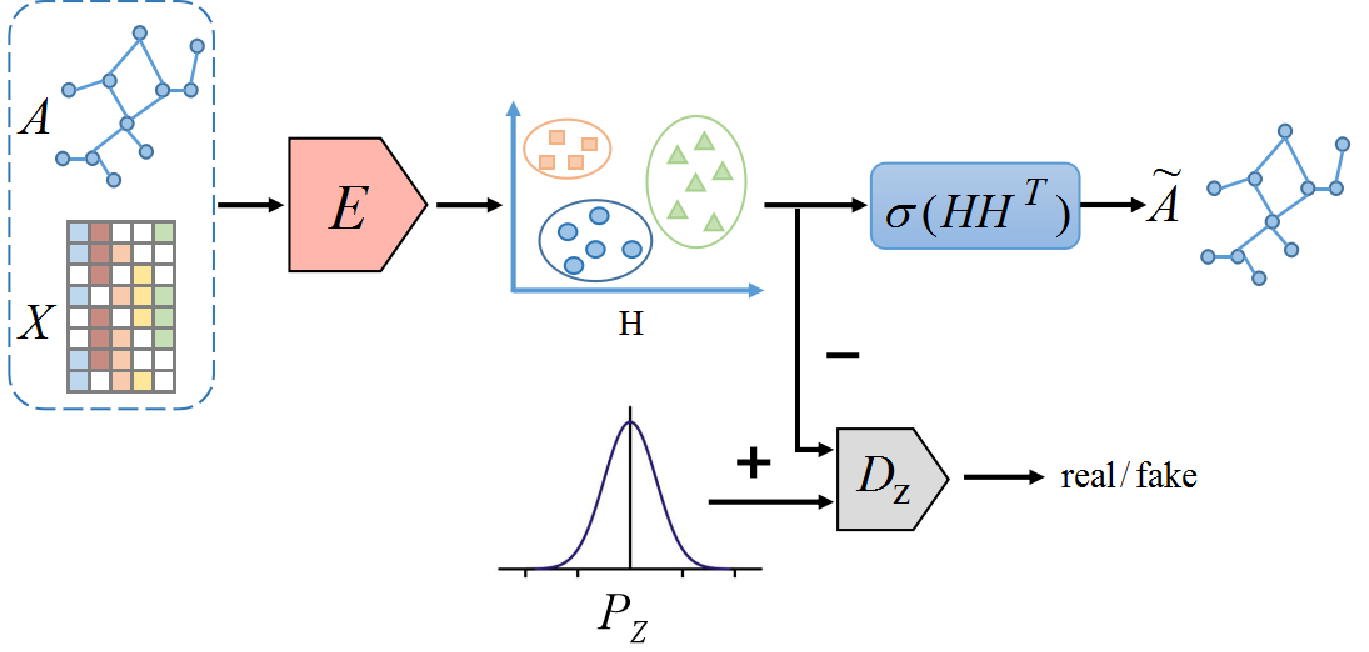}
	\caption{architecture of ARGA\cite{ARGA} and AIDW\cite{AIDW}. $A$ and $\tilde{A}$ represent the adjacency matrix and reconstructed adjacency matrix, respectively. $X$ denotes the node feature matrix. $"+"$ denotes the real samples and $"-"$ denotes the fake samples.}
	\label{ARGA}
\end{figure}

GAN\cite{GAN} has demonstrated its strong distribution fitting ability in various fields since it was first proposed by Goodfellow. AAE\cite{AAE}, BiGAN\cite{BiGAN}, and ALI\cite{ALI} have already explored the application of adversarial learning in the field of image representation. 
And most recently, BigBiGAN\cite{BigBiGAN} based on BiGAN has achieved amazing performance in image representation learning. 

The success of the above works shows that the distribution fitting ability of GAN can be used not only to generate data but also to understand data. Thus, GAN has been introduced into the field of graph representation learning in various forms\cite{ARGA,AIDW,graphgan}. \textbf{From the perspective of sample generation}, Ding et al.\cite{GraphSGAN} use the generator to generate fake samples in low-density areas between subgraphs to enable the classifier to take into account the density characteristics of the graph data. 
To preserve the structure information, ProGAN\cite{ProGAN} applies the generator to generate triplets of nodes to discover the proximity in the original space and preserving it in the low dimensional space. \textbf{From the perspective of latent distribution fitting}, NetRA\cite{NetRA} uses adversarial learning to keep the latent representations away from the noise representation generated by the normal distribution to improve the anti-jamming capability of the representations. As shown in Fig.\ref{ARGA}, ARGA\cite{ARGA} and AIDW\cite{AIDW} take a similar approach and introduce adversarial learning into \cite{GAE} and \cite{deepwalk} respectively, to improve the generalization ability of the representations.

\begin{figure*}[t]	
	\centering 
	\includegraphics[width=6.4in]{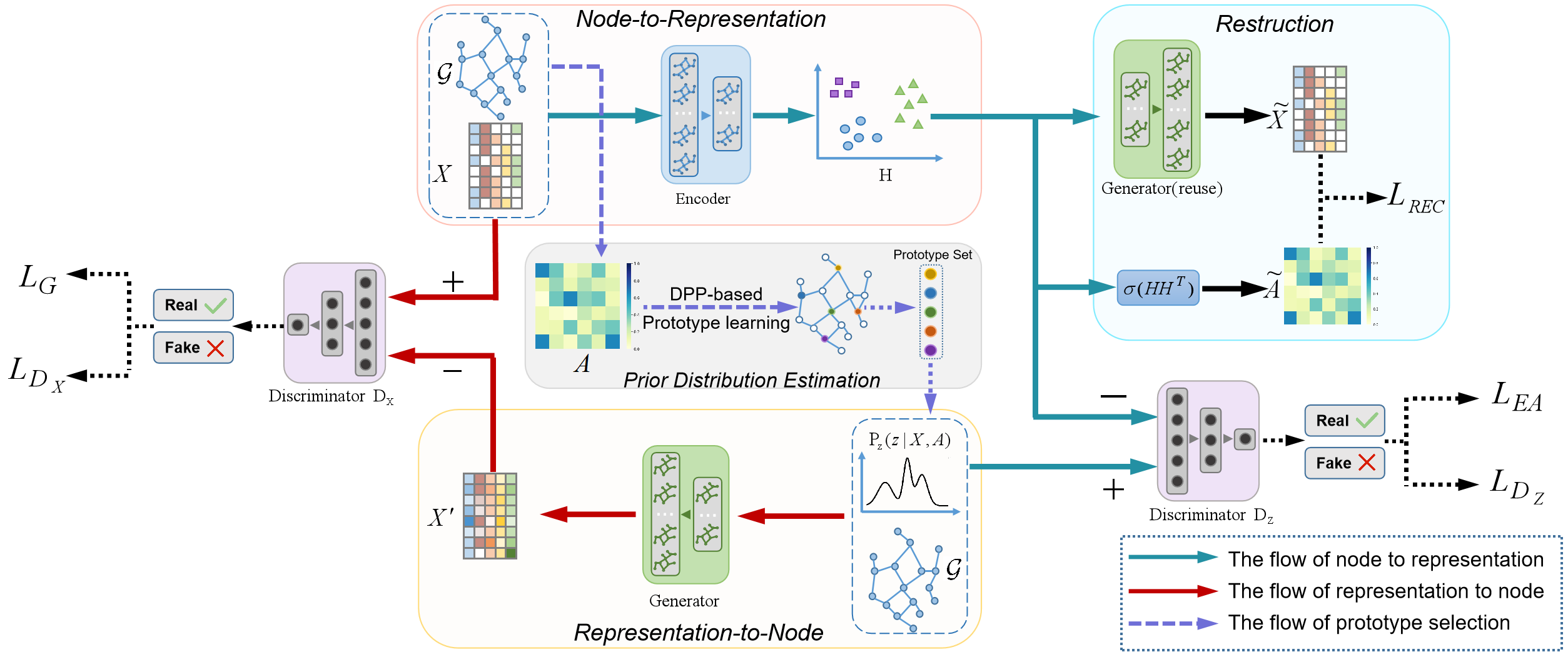}
	\caption{Architecture overview of our DBGAN. $A$ and $\tilde{A}$ represent the adjacency and reconstructed adjacency matrix, respectively. $X$, ${X}'$, and $\tilde{X}$ denote the node raw feature, the generated feature, and the reconstructed feature, respectively. $L_{REC}$ denotes reconstruction loss, $L_{G}, L_{EA}, L_{D_{X}}$, and $L_{D_{Z}}$ denote the adversarial loss for $G, E, D_{X}$, and $D_{Z}$, respectively. And $\mathbb{P}_{z}(z|X,A)$ denotes the estimated prior distribution,  $``+``$ and$``-"$ represent the real and fake samples, respectively.}
	\label{DBGAN}
	
\end{figure*}
Although the above methods have achieved good performance, the disadvantages of them are also obvious. \cite{ProGAN, GraphSGAN, graphgan} only consider the local structure information, ignores the global structure and distribution consistency, resulting in noise sensitivity, which makes the representation less robust. Besides, in \cite{AIDW, ARGA}, the node feature hasn't been utilized, and the pre-assumed normal distribution won't ideally conform to the complex graph data in reality, which makes the model tend to be over-smoothing and further reduces the generalization of the learned latent representation.

\section{Methodology}

In this section, we first give the problem definition of graph representation learning, then present bidirectional adversarial learning in DBGAN, and finally introduce a prior distribution estimation method for latent representation by prototype learning.

\subsection{Problem Definition}
An undirected graph is given as $\mathcal{G} = (V,\mathcal{E})$, where $V = \{{v_{1},\cdots,v_{n}}\}$ consists of a set of nodes with $|V| = n$, and $\mathcal{E}$ is a sets of edges with $e_{ij}\in \mathcal{E}$. $X=\{x_{1},\cdots,x_{n}\} \in \mathbb{R}^{d \times n}$ denotes the node feature matrix of a graph, where $x_{i}$ represents the raw feature of node $v_{i}$. The graph structure can be represented by the adjacency matrix $A$ with $A_{ij}$ = 1 if $e_{ij} \in \mathcal{E}$, otherwise $A_{ij}$= 0. The degree matrix is represented by diagonal matrix $D$ with $D_{ii} = \sum {_{j}}A_{ij}$, and $D^{-\frac{1}{2}}AD^{-\frac{1}{2}}$ is the normalized adjacency matrix. In the following sections, we denote $A$ as the normalized adjacency matrix.

For a given graph $\mathcal{G}$, graph representation learning aims to map nodes $v_{i} \in V$ to latent representation $h_{i} \in H$ where $H = \{h_{1},\cdots,h_{n}\} \in \mathbb{R}^{q \times n}$ denotes the latent representation matrix. In particular, both the structure of $A$ and the node content from $X$ are expected to be well preserved in $H$ space.
\subsection{Overall Framework}  The overall framework of DBGAN is shown in Fig.\ref{DBGAN}. In the encoding phase, the encoder $E$ accepts $A$ and $X$ as inputs and outputs a latent representation matrix $H$. After that, $E$ and the data $Z$ sampled from the prior distribution $P_{z|(X,A)}$ are input into the discriminator $D_{z}$ for adversarial training, where $z_{i} \in Z$ and $h_{i} \in H$ are positive and negative samples, respectively. Meanwhile, the generator $G$ accepts $Z$ and $A$ as input and outputs the fake feature matrix ${X}'$ of the graph, after which ${X}'$ as negative samples and $X$ as positive samples are sent to the discriminator $D_{x}$ for adversarial training. In the reconstruction phase, $H$ is fed to $G$, and then outputs the rebuilt $\tilde{X}$. In addition, $H$ is reconstructed into $\tilde{A}$ through the reconstruction process $\tilde{A} = sigmoid(HH^{T})$. In this work, we use GCN\cite{GCN} as encoder $E$ and generator $G$, and MLP for both discriminators $D_{z}$ and $D_{x}$.
\subsection{Bidirectional Adversarial Learning}
Different from adversarial learning as in AIDW\cite{AIDW} and ARGA\cite{ARGA}, we propose a bidirectional adversarial learning algorithm that establishes a mutual mapping between graph data and latent representation. It is capable of balancing the consistency between distribution-level and sample-level, thus leading to a significant improvement of generalization ability in latent representation space.

The bidirectional adversarial learning is mainly implemented in two streams. One is composed of $E$ and $D_{z}$ to model the mapping from graph data to representation, and the other is composed of $G$ and $D_{x}$ for the reverse mapping. 
Completely different from the bidirectional adversarial learning in \cite{BiGAN}, our DBGAN makes full use of the prior distribution in latent space, which acts as not only the target of output for the encoder $E$ but also the source of input for the generator $G$.
The superiority of our bidirectional adversarial learning method can be claimed in three aspects: \textbf{(i)} bidirectional mapping is more beneficial to exploiting the inherent graph structure than unidirectional mapping. It facilitates the trade-off of distribution-level consistency and sample-level consistency, resulting in more generalized representations; \textbf{(ii)} the application of adversarial learning  in our DBGAN can address the over-fitting problem well, which to some extent improves the robustness of representation;
\textbf{(iii)} if the capacity is allowed to be sufficient for encoder and decoder, the auto-encoder may degrade into a copying task instead of extracting more useful information about the data distribution~\cite{deeplearning}. However, the capacity has no effect on DBGAN, since $G$ and $E$ will not be optimized synchronously with the same batch of data, thus enforcing the reconstruction constraints on latent representation.
\vspace{-0.2cm}
\paragraph{Adversarial loss.} Adversarial loss is used to minimize the distance between two distributions. Here we use the Wasserstein distance in \cite{WGAN} to measure the difference between the graph data distribution $\mathbb{P}_{r}(x)$ and prior distribution of latent representation $\mathbb{P}_{z}(x)$, and it can be defined as 
\begin{equation}
W\left [ \mathbb{P}_{z},\mathbb{P}_{r}\right ] = \max \limits_{f,\left \| f \right \|_L\leq 1} \mathbb{E}_{z \sim \mathbb{P}_{z}} \left[ f \left( z \right) \right] - \mathbb{E}_{x \sim \mathbb{P}_{r}}\left[ f\left( E(x)\right) \right]
\label{w_loss}
\end{equation}
where $f$ denotes the discriminant function, and $\left \| f \right \|_L\leq 1$ represents a condition that the discriminant function needs to satisfy the Lipschitz constraint with Lipschitz constant 1.
Here the gradient penalty term proposed in \cite{WGAN-GP} is used to implement the Lipschitz constraint and the discriminant function is learned by the discriminator $D_{z}$. Hence, Eq.(\ref{w_loss}) can be taken as the objective of $D_{z}$, while the objective of $E$ is the opposite. According to Eq.(\ref{w_loss}), we can define the adversarial losses of $D_{z}$ and $E$ as follows
\begin{align}
\mathcal{L}_{D_{z}}(z,x) = -&\mathbb{E}_{z\sim \mathbb{P}_{z}}[D_{z}(z)]+\mathbb{E}_{x\sim \mathbb{P}_{r}}[D_{z}(E(x))] \nonumber\\+ &\lambda \mathbb{E}_{\hat{z}\sim \mathbb{P}_{\hat{z}}}[\left \| \bigtriangledown _{\hat{z}}D_{z}({\hat{z}})-1 \right \|] \\
\mathcal{L}_{EA}(z,x) = &\mathbb{E}_{z\sim \mathbb{P}_{z}}[D_{z}(z)]-\mathbb{E}_{x\sim \mathbb{P}_{r}}[D_{z}(E(x))]
\label{E_loss}
\end{align}
where $\hat{z}$ denotes random interpolation of $E(x)$ and $z$ sampled from $\mathbb{P}_{z}$.
When $E$ is updated, $D_{z}$ will not change. Thus, $\mathbb{E}_{z\sim \mathbb{P}_{z}}[D_{z}(z)]$ in Eq.(\ref{E_loss}) will not provide gradients for $E$, and then Eq.(\ref{E_loss}) can be simplified as
\begin{align}
\mathcal{L}_{EA}(x) = -\mathbb{E}_{x\sim \mathbb{P}_{r}}[D_{z}(E(x))]
\end{align}
Likewise, by switching the roles of $\mathbb{P}_{z}$ and $\mathbb{P}_{x}$, we can get the adversarial losses of $G$ and $D_{x}$ as follows
\begin{align}
\mathcal{L}_{D_{x}}(x,z) = -&\mathbb{E}_{x\sim \mathbb{P}_{r}}[D_{x}(x)]+\mathbb{E}_{z\sim \mathbb{P}_{z}}[D_{x}(G(z))] \nonumber\\+ &\lambda \mathbb{E}_{\hat{x}\sim \mathbb{P}_{\hat{x}}}[\left \| \bigtriangledown _{\hat{x}}D_{x}({\hat{x}})-1 \right \|] \\
\mathcal{L}_{G}(z) = -&\mathbb{E}_{z\sim \mathbb{P}_{z}}[D_{x}(G(z))]
\label{G_loss}
\end{align}
where $\hat{x}$ denotes random interpolation of $G(z)$ and $x$ sampled from $\mathbb{P}_{x}$.
\vspace{-0.4cm}
\paragraph{Reconstruction loss.} In addition to the adversarial loss that guarantees the distribution-level consistency between the graph space and raw feature space, the reconstruction loss $\mathcal{L}_{REC}(x)$ is also enforced for sample-level consistency. This is essential to further improve the representation ability in latent representation space, by both node feature reconstruction and adjacency matrix reconstruction.

We follow the settings in \cite{GAE} to get the reconstructed adjacency matrix $\tilde{A}$ from the latent representation, and here $\tilde{A}$ should be similar to real adjacency matrix $A$. Besides, by the mapping established by $G$ of the latent representations to the graph data, we can get the reconstructed feature matrix ${X}' = G(E(X))$. The reconstruction loss can be defined as follows
\begin{align}
\mathcal{L}_{REC}(x) = \mathbb{E}_{x\sim \mathbb{P}_{r}}[d(X,{X}')] + \mathbb{E}_{x\sim \mathbb{P}_{r}}[d(A,\tilde{A})]
\label{rec_loss}
\end{align}
where ${X}' = G(E(X))$, ${A}' = sigmoid(E(X)\cdot E(X)^{T})$, and $d(x,y) = x\log y +(1-x)\log (1-y)$. Therefore, the overall loss of the encoder $E$ can be written as
\begin{align}
\mathcal{L}_{E}(x) = \mathcal{L}_{EA}(x) + \alpha \mathcal{L}_{REC}(x)
\label{E_loss_final}
\end{align}

\textcolor{black}{
	It is worth noting that the effectiveness of our DBGAN can be claimed by Theorem 1.
}

\vspace{-0.3cm}
\paragraph{Theorem 1.}
\textit{Assuming $W\left [ \mathbb{P}_{z},\mathbb{P}_{r}\right ]$ and $W\left [ \mathbb{P}_{r},\mathbb{P}_{z}\right ]$ converge, i.e., $H = E(X) \sim \mathbb{P}_{z}$, and $G(Z) \sim \mathbb{P}_{r}$, it can be inferred that ${X}' = G(E(X)) \sim \mathbb{P}_{r}$. Thus, ${X}'$ and $X$ will obey an identical distribution $\mathbb{P}_{r}$, i.e., ${X}' \sim \mathbb{P}_{r} $ and $X \sim \mathbb{P}_{r} $. Then, $X \approx G(E(X))$ can be obtained as the reconstruction error converges.  }

%

\subsection{Prior distribution estimation for latent representation}
For the methods based on prior distribution assumptions\cite{ARGA,AIDW}, the prior distribution $\mathbb{P}_{z}$ is critical to their performances. For example, for graph data with multiple categories, it is not reasonable to use normal distribution $\mathbb{N}(0,1)$ as $\mathbb{P}_{z}$ to represent the graph. Besides, by bidirectional adversarial learning, an appropriate $\mathbb{P}_{z}$ can improve the robustness and discriminability of the representation. Since we have no more priors except for the given $A$ and $X$, an intuitive approach is to estimate $\mathbb{P}_{z}(z|X)$ that approximates to $\mathbb{P}_{z}(z)$ by a non-parametric estimation method such as Kernel Density Estimation (KDE). In addition, we use PCA to reduce the dimension of $X$ to get $X_{p} = \{x_{i}\}_{i=1,\cdots,n}$, and then we can get $\mathbb{P}_{z}(z|X)$ as follows
\begin{align}
\mathbb{P}_{z}(z|X) =\frac{1}{n}\sum_{i=1}^{n}K_{b}(z-x_{i})=\frac{1}{nb}\sum_{i=1}^{n}K(\frac{z-x_{i}}{b})
\label{kde_pzx}
\end{align}
where $K(\cdot)$ is a kernel function, $b$ denotes the bandwidth, and $K_{b}(\cdot)$ is the scaled kernel function.

However, there are some problems with this intuitive approach. First, the explicit structural information embedded in $A$ is completely ignored; second, the learned model is susceptible to the noisy $X$, thus reducing the robustness of representation. Therefore, 
we can approximate $\mathbb{P}_{z}(z)$ using $\mathbb{P}_{z}(z|X,A)$ instead of $\mathbb{P}_{z}(z|X)$.
\vspace{-0.3cm}
\paragraph{DPP-based prototype learning.}
For heterogeneous $A$ and $X$, it is not trivial to obtain $\mathbb{P}_{z}(z|X,A)$ directly. Considering that $A$ and $X$ are structurally consistent though they are in different domains, we can utilize the cross-domain prototypes to bridge the raw feature domain and the graph domain. Thus $\mathbb{P}_{z}(z|X,A)$ can be replaced with $\mathbb{P}_{z}(z|X_{S_{p}},A_{S_{p}})$, where $S_{p}$ denotes the index set for prototypes.

For prototype learning, the Determinant Point Process (DPP)\cite{kdpp} is adopted to select a diversified prototype subset. Specifically, the adjacency matrix $A$ is considered as the measure matrix. Given a subset $V_{S}\subseteq V$, whose items are indexed by $S\subseteq \mathcal{N} = \{1,\cdots,n\}$, then the sampling probability of $S$ based on the measure matrix $A$ can be defined as follows
\begin{align}
P_{A}(S)=\frac{\det(A_{S})}{ \det(A+I)}
\label{IP}
\end{align}
where $I$ denotes the identity matrix, $ A_{S} \equiv [A_{ij}]_{i,j \in S} $, and $\det(\cdot)$ denotes the determinant of a matrix. Obviously, sampling probability defined here is normalized because of
\begin{align}
\sum_{S\subseteq N}\det(A_{S})=\det(A+I)
\label{IP=1}
\end{align}
According to Eq.(\ref{IP}), a probability will be assigned to any subset of $\mathcal{N}$, which will result in a large search range for the prototype index subset. Hence, we have limited the subset size to $|S|=m$. When the size of subset $S$ is fixed to $m$, we can define the sampling probability as follows
\begin{align}
P_{A}^{k}(S)=\frac{\det(A_{S})}{ \sum_{\left |{S}'  \right |=k}\det(A_{{S}'})}
\label{IPk}
\end{align}
Similarly, according to Eq.(\ref{IP=1}), $P_{A}^{k}(S)$ is also normalized. 

We explain the definition of importance probability from the geometric explanation of the matrix determinant. Considering $A_{ij}$ is computed from $\varphi(v_{i})$ and $\varphi(v_{j})$, where $\varphi(\cdot)$ is a nonlinear mapping function, then $\det(A)$ can be interpreted as the volume of the geometry spanned by the nodes $v_{i} \in V$ \cite{kdpp}. 
Therefore, the prototypes $S_{p}$ measured by $P_{A}^{k}(S_{p})$ can better sketch the consistent distribution of $A$ and $X$.
\vspace{-0.4cm}
\paragraph{Structure-aware prior distribution estimation.}
According to the prototype index set $S_{p}$ with $|S_{p}| = m$, a node feature matrix $X_{p}$ can be sampled from $X$. Then, we use PCA to reduce the dimension of $X_{p}$ to get $H_{p}$. Assuming $h_{i} \in H_{p}$ is i.i.d., $\mathbb{P}_{z}(z|X_{S_{p}},A_{S_{p}})$ can be defined by
\begin{align}
\mathbb{P}_{z}(z|X_{S_{p}},A_{S_{p}}) =\frac{1}{m}\sum_{i=1}^{m}K_{b}(z-h_{i})=\frac{1}{mb}\sum_{i=1}^{m}K(\frac{z-h_{i}}{b})
\label{kde}
\end{align}

In summary, with the flow in Eq.\ref{flow}, we obtain the approximation of $P_{z}$, i.e., $\mathbb{P}_{z}(z|X_{S_{p}},A_{S_{p}})$.
\begin{align}
\mathbb{P}_{z}(z) \rightarrow \mathbb{P}_{z}(z|X)\rightarrow \mathbb{P}_{z}(z|X,A) \rightarrow \mathbb{P}_{z}(z|X_{S_{p}},A_{S_{p}})
\label{flow}
\end{align}

\section{Experimental Results and Analysis}
We first detail our experimental protocol, and then present comparison results of DBGAN with the state of the art for graph representation learning.

\subsection{Evaluation Setup and Metrics}
\paragraph{Datasets.} We select three widely used graph datasets, Cora\cite{Cora}, Citeseer\cite{Citeseer}, and Pubmed\cite{Pubmed}, to verify the performance of DBGAN in unsupervised representation learning. Each dataset contains a complete node feature matrix $X$ and an adjacency matrix $A$. Details of three dataset statistics are in Table~\ref{datasets}.

\begin{table}[t]
	\centering
	\caption{Statistics of the used datasets.}
	\label{datasets}

	\begin{tabular}{@{}c|c|c|c|c@{}}
		\toprule
		\textbf{Dataset}  & \textbf{\#Nodes} &  \textbf{\#Edges} & \textbf{\#Classes} &  \textbf{\#Features} \\ \midrule
		Cora     & 2708  & 5429  & 7       & 1433     \\
		Citeseer & 3327  & 4732  & 6       & 3703     \\
		Pubmed   & 19717 & 44338 & 3       & 500      \\ \bottomrule
	\end{tabular}
\vspace{-0.5cm}
\end{table}

\begin{table*}[t]
	\caption{Experimental results of link prediction.}
	\normalsize
	\centering
	\begin{tabular}{@{}c|cc|cc|cc@{}}
		\toprule
		\multirow{2}{*}{\textbf{Methods}} & \multicolumn{2}{c|}{\textbf{Cora}}       & \multicolumn{2}{c|}{\textbf{Citeseer}}   & \multicolumn{2}{c}{\textbf{Pubmed}}    \\ \cmidrule(l){2-7} 
		& \textbf{AUC}            & \textbf{AP}             & \textbf{AUC}            & \textbf{AP}             & \textbf{AUC}            & \textbf{AP}            \\ \midrule
		Spectral\cite{spectral}                       & 84.6$\pm$0.01  & 88.5$\pm$0.00  & 80.5$\pm$0.01  & 85.0$\pm$0.01  & 84.2$\pm$0.02  & 87.8$\pm$0.01 \\
		DeepWalk \cite{deepwalk}                      & 83.1$\pm$0.01  & 85.0$\pm$0.00  & 80.5$\pm$0.01  & 83.6$\pm$0.01  & 84.4$\pm$0.00  & 84.1$\pm$0.00 \\ \midrule
		GAE\cite{GAE}                      & 91.0$\pm$0.02  & 92.0$\pm$0.03  & 89.5$\pm$0.04  & 89.9$\pm$0.05  & 96.4$\pm$0.00  & 96.5$\pm$0.00 \\
		VGAE\cite{GAE}                     & 91.4$\pm$0.01  & 92.6$\pm$0.01  & 90.8$\pm$0.02  & 92.0$\pm$0.02  & 94.4$\pm$0.02  & 94.7$\pm$0.02 \\
		ARGA\cite{ARGA}                     & 92.4$\pm$0.003 & {\color{blue}93.2$\pm$0.003 } & 91.9$\pm$0.003 & 93.0$\pm$0.003 & {\color{blue}96.8$\pm$0.001 } & {\color{blue}97.1$\pm$0.01 } \\
		ARVGA \cite{ARGA}                   & 92.4$\pm$0.004 & 92.6$\pm$0.004 & 92.4$\pm$0.003 & 93.0$\pm$0.03  & 96.5$\pm$0.001 & 96.8$\pm$0.01 \\
		DGI\cite{DGI}                      & {\color{blue} 92.6$\pm$0.02} & 93.1$\pm$0.01 & 93.3$\pm$0.04 & 94.1$\pm$0.03  & 95.9$\pm$0.002 & 96.3$\pm$0.01 \\
		GALA\cite{GALA}                     & -            & -            & {\color{blue}94.4$\pm$0.009 }             & {\color{blue}94.8$\pm$0.01 }             & -             & -            \\ \midrule
		DBGAN                    & {\color{red} 94.5$\pm$0.01}            & {\color{red} 95.1$\pm$0.05 }           & {\color{red}94.5$\pm$0.04 }             & {\color{red}95.8$\pm$0.01 }             & {\color{red}96.8$\pm$0.01 }             & {\color{red}97.3$\pm$0.02 }            \\ \bottomrule
	\end{tabular}
	\vspace{-0.3cm}
	\label{link prediction}
\end{table*}
\vspace{-0.5cm}
\paragraph{Protocols and evaluation metrics.}
The tasks of link prediction and node clustering are employed to evaluate the discrimination and generalization of learned node representation. 
In particular, for link prediction, we divided each dataset into a training set, a test set, and a validation set, with a ratio of 85:5:10. 
To avoid the influence of randomness, we average the results over 20 times of execution with different training set selections as in \cite{GAE}. 
Then the mean scores and standard errors of Area Under Curve (AUC) and Average Precision (AP) are reported.
While for node clustering, we adopt Kmeans~\cite{Kmeans} to classify the learned representations into several clusters.
As in \cite{GALA}, accuracy (ACC), normalized mutual information (NMI), and adjusted rand index (ARI) are used to measure the performance of clustering. 
Likewise, we still report the averaged results over 20 times of execution.

\vspace{-0.5cm}
\paragraph{Implementation details.}
For the flow from latent representation to node as in Fig. \ref{DBGAN}, we follow the training strategy in WGAN-GP~\cite{WGAN-GP}, where a complete iterative process is to train $G$ once after training $D_{x}$ 5 times.
In addition, the discriminator and encoder in our DBGAN are trained \textcolor{black}{synchronously}, since encoder $E$ is optimized for both reconstruction loss and adversarial loss.
The model uses \textcolor{black}{Adam~\cite{Adam}} as the optimizer with $\beta_{1}=0.9$ and $\beta_{2}=0.999$, and is implemented on the Tensorflow platform.
\vspace{-0.5cm}

\paragraph{Comparison methods.}
We choose to compare with a total of fifteen unsupervised graph representation algorithms, especially those that have achieved the state-of-the-art results recently. In particular, such compared algorithms can be divided into three groups.

\begin{itemize}
	\item[\uppercase\expandafter{\romannumeral1.}]{ \emph {Using node feature or graph structure only.} }
	In general, Kmeans\cite{Kmeans} is considered as a baseline for node clustering.
	Due to merely usage of topological structure of the graph, Spectral Clustering\cite{spectral} usually serves as a typical social network representation learning algorithm. 
	Big-Clam\cite{Bigclam} is a large-scale community detection algorithm based on non-negative matrix factorization.
	Additionally, as one of the most representative graph representation learning algorithms, we compare with DeepWalk\cite{deepwalk} which encodes graph nodes into latent representations by random walks.
	A recent algorithm DNGR\cite{DNGR} using auto-encoder to preserve graph structure is also employed.
	
	\item[\uppercase\expandafter{\romannumeral2.}]{\emph{Using both node feature and graph structure.}}
	Circles\cite{Circles} is a node clustering algorithm that treats each node as ego and builds an ego graph that preserves the original connection relationship.
	RTM\cite{RTM} aims to learn topic distributions of each document from text.
	RMSC\cite{RMSC} is a multi-view clustering algorithm that can effectively remove noise.
	TADW\cite{TADW} integrates node content into Deepwalk, and explains Deepwalk by matrix factorization.
	
	\item[\uppercase\expandafter{\romannumeral3.}]{\emph{Using node feature and graph structure both with GCN.}}
	GAE\cite{GAE} is the first GCN-based auto-encoder algorithm for unsupervised graph representation learning.
	VGAE\cite{GAE} is a variational version of GAE.
	ARGA\cite{ARGA} is another variant of GAE that introduces adversarial learning into GAE. Similarly, VARGA\cite{ARGA} is a variational version of ARGA.
	DGI\cite{DGI} is a GCN-based method which generates node representations by maximizing local mutual information in the patch representation of the graph.
	GALA\cite{GALA} is the latest GCN-based unsupervised framework for graph data, which designs a decoder with Laplacian sharpening as an improvement of GAE.
\end{itemize}

\begin{table*}[]
	\centering
	\caption{Experimental results of node clustering.}
	\resizebox{\textwidth}{!}{
	\begin{tabular}{l|l|ccc|ccc|ccc}
		\hline
		\multicolumn{2}{c|}{\multirow{2}{*}{\textbf{Methods}}}                   & \multicolumn{3}{c|}{\textbf{Cora}}                                                                                                   & \multicolumn{3}{c|}{\textbf{Citeseer}}                                                                                            & \multicolumn{3}{c}{\textbf{Pubmed}}                                                                                              \\ \cline{3-11} 
		\multicolumn{2}{c|}{}                                           & \textbf{ACC}           & \textbf{NMI}           & \textbf{ARI}           & \textbf{ACC}          & \textbf{NMI}          & \textbf{ARI}          & \textbf{ACC}          & \textbf{NMI}          & \textbf{ARI}          \\ \hline
		\multirow{5}{*}{\uppercase\expandafter{\romannumeral1}} & Kmeans\cite{Kmeans}     & 0.492                                   & 0.321                                   & 0.229                                   & 0.540                                  & 0.305                                  & 0.278                                  & 0.595                                  & 0.315                                  & 0.281                                  \\ \cline{2-11} 
		& Spectral\cite{spectral} & 0.367                                   & 0.126                                   & 0.031                                   & 0.238                                  & 0.055                                  & 0.010                                  & 0.528                                  & 0.097                                  & 0.062                                  \\
		& Big-Clam\cite{Bigclam}  & 0.271                                   & 0.007                                   & 0.001                                   & 0.250                                  & 0.037                                  & 0.007                                  & 0.394                                  & 0.006                                  & 0.003                                  \\
		& DeepWalk\cite{deepwalk} & 0.484                                   & 0.327                                   & 0.242                                   & 0.336                                  & 0.087                                  & 0.092                                  & 0.684                                  & 0.279                                  & 0.299                                  \\
		& DNGR\cite{DNGR}         & 0.419                                   & 0.318                                   & 0.142                                   & 0.325                                  & 0.180                                  & 0.042                                  & 0.458                                  & 0.155                                  & 0.054                                  \\ \hline
		\multirow{4}{*}{\uppercase\expandafter{\romannumeral2}} & Circles\cite{Circles}   & 0.606                                   & 0.404                                   & 0.362                                   & 0.571                                  & 0.300                                  & 0.293                                  & -                                      & -                                      & -                                      \\
		& RTM\cite{RTM}           & 0.439                                   & 0.230                                   & 0.169                                   & 0.450                                  & 0.239                                  & 0.202                                  & 0.574                                  & 0.194                                  & 0.444                                  \\
		& RMSC\cite{RMSC}         & 0.406                                   & 0.255                                   & 0.089                                   & 0.295                                  & 0.138                                  & 0.048                                  & 0.576                                  & 0.255                                  & 0.222                                  \\
		& TADW\cite{TADW}         & 0.560                                   & 0.441                                   & 0.332                                   & 0.454                                  & 0.291                                  & 0.228                                  & -                                      & -                                      & -                                      \\ \hline
		\multirow{6}{*}{\uppercase\expandafter{\romannumeral3}} & GAE\cite{GAE}           & 0.596                                   & 0.429                                   & 0.347                                   & 0.408                                  & 0.176                                  & 0.124                                  & 0.672                                  & 0.277                                  & 0.279                                  \\
		& VGAE\cite{GAE}          & 0.502                                   & 0.329                                   & 0.254                                   & 0.467                                  & 0.260                                  & 0.205                                  & 0.630                                  & 0.229                                  & 0.213                                  \\
		& ARGA\cite{ARGA}         & 0.640                                   & 0.449                                   & 0.352                                   & 0.573                                  & 0.350                                  & 0.341                                  & 0.668                                  & 0.305                                  & 0.295                                  \\
		& ARVGA\cite{ARGA}        & 0.638                                   & 0.450                                   & 0.374                                   & 0.544                                  & 0.261                                  & 0.245                                  & 0.690                                  & 0.290                                  & 0.306                                  \\
		& DGI\cite{DGI}           & 0.554                                   & 0.411                                   & 0.327                                   & 0.514                                  & 0.315                                  & 0.326                                  & 0.589                                  & 0.277                                  & 0.315                                  \\
		& GALA\cite{GALA}         & {\color{blue} 0.745}  & {\color{red} 0.576} & {\color{blue} 0.531} & {\color{red}0.693}   & {\color{red}0.441}   & {\color{red}0.446}  & {\color{blue}0.693}  & {\color{red}0.327}   & {\color{blue}0.321} \\ \cline{2-11} 
		& DBGAN                                    & {\color{red} 0.748$\pm$0.005}  & {\color{blue} 0.560$\pm$0.005} & {\color{red} 0.540$\pm$0.01} & {\color{blue}0.670$\pm$0.006}         & {\color{blue}0.407$\pm$0.005}        & {\color{blue}0.414$\pm$0.01}        & {\color{red}0.694$\pm$0.001}        & {\color{blue}0.324$\pm$0.07}        & {\color{red}0.327$\pm$0.02}  \\ \hline
	\end{tabular}
}
	\label{node cluster}
\end{table*}

\subsection{Evaluation on Link Prediction}
For link prediction task, the hyperparameters $\alpha$ and $\lambda$ are set to 1 on all three datasets, the two hidden layers of $G$ are set to 256-neuron and 512-neuron respectively, and the two hidden layers of $D_{x}$ are set to 512-neuron and 256-neuron respectively. In particular, on Cora dataset, we set the hidden and output layers of $E$ to 32-neuron, and the two hidden layers of $D_{z}$ are set to 64-neuron and 32-neuron respectively. While on Citeseer and Pubmed datasets, we set the hidden layer and output layer of $E$ to 64-neuron, and the two hidden layers of $D_{z}$ also to 64-neuron. 

The comparative results on link prediction task are shown in Table \ref{link prediction}. 
It can be concluded that, (i) compared to Spectral Clustering and DeepWalk, the spectral convolution-based auto-encoder framework can effectively improve the performance of graph representation in connection prediction tasks. In particular, our DBGAN has achieved the best performance on all three datasets, with an improvement of $0.1\%\sim 1.9\%$ w.r.t. AUC, and $0.2\%\sim 1.9\%$ w.r.t. AP, over the strongest competitor;
(ii) our DBGAN outperforms ARGA that also introduces adversarial learning by about 2.0\% and 2.5\% on Cora and Citeseer datasets respectively. This just verifies that the effectiveness of our proposed structure-aware prior distribution estimation by DPP-based prototype learning;
(iii) the approximate 1.0\% improvement of our DBGAN on Citeseer dataset over GALA that is the state-of-the-art method is achieved. This means that our DBGAN has created a new baseline in the area of graph representation learning.
\subsection{Evaluation on Node Clustering}
For node clustering task, the network setup for $G$ and $D_{x}$ are the same as those for link prediction task. In particular, on Cora dataset, we set the hidden and output layers of $E$ to 64-neuron and 128-neuron, the two hidden layers of $D_{z}$ to 64-neuron, and the hyperparameters $\alpha$ and $\lambda$ to 0.01 and 1 respectively. On Citeseer and Pubmed datasets, we set the hidden layer and output layer of $E$ to 64-neuron, the two hidden layers of $D_{z}$ also to 64-neuron, and the hyperparameters $\alpha$ and $\lambda$ to 1e-5 and 1 respectively. 

We present the comparative results on node clustering in Table \ref{node cluster}.
It can be observed that, (i) the performance of such algorithms that use both node features and graph structure can outperform significantly than those using only one of them;
(ii) Kmeans~\cite{Kmeans} that uses only node features improves the overall performance of the methods only using graph structures, by an obvious margin (from 0.80\% to 20.4\% for ACC). This validates that introducing node features is necessary for node clustering tasks;
(iii) it is worth noting that, although GALA\cite{GALA} outperforms our DBGAN slightly in several cases, it is indeed at the cost of high dimension of latent representation (e.g., 400 for GALA and 128 for ours on Cora, and 500 and 64 on Citeseer). 

\begin{figure*}[t]
	\centering
	\subfigure[Citeseer(raw)]{
		\centering
		\includegraphics[width=1.6in]{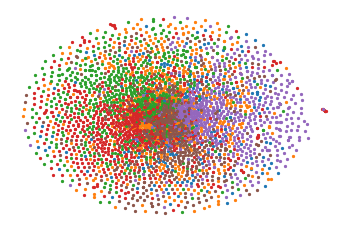}
	}%
	\subfigure[Citeseer(GAE)]{
		\centering
		\includegraphics[width=1.6in]{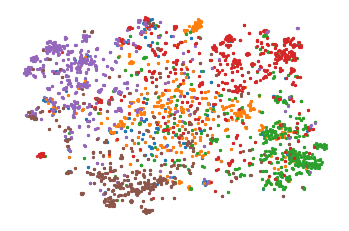}
	}%
	\subfigure[Citeseer(DGI)]{
		\centering
		\includegraphics[width=1.6in]{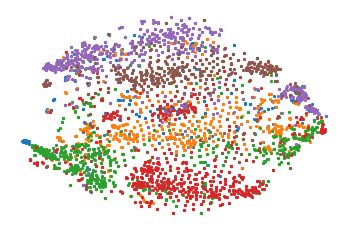}
	}%
	\subfigure[Citeseer(DBGAN)]{
		\centering
		\includegraphics[width=1.6in]{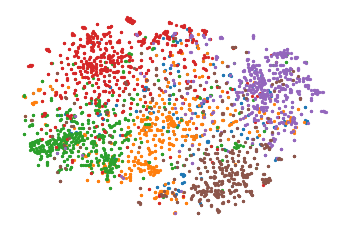}
	}%
	\centering
	\caption{Visualization of the Citeseer dataset.}.
	\label{fig2}
	\vspace{-0.5cm}
\end{figure*}

\begin{figure*}[t]
	\centering
	\subfigure[Cora(raw)]{
		\centering
		\includegraphics[width=1.6in]{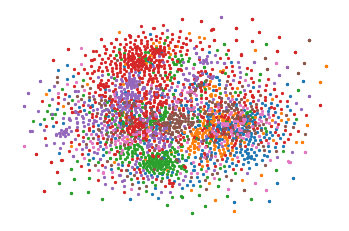}
	}%
	\subfigure[Cora(GAE)]{
		\centering
		\includegraphics[width=1.6in]{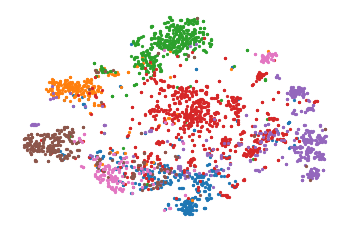}
	}%
	\subfigure[Cora(DGI)]{
		\centering
		\includegraphics[width=1.6in]{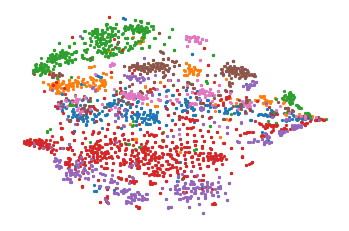}
	}%
	\subfigure[Cora(DBGAN)]{
		\centering
		\includegraphics[width=1.6in]{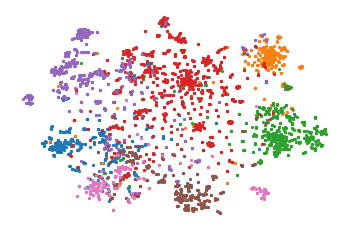}
	}%
	\centering
	\caption{Visualization of the Cora dataset.}.
	\label{fig3}
	\vspace{-0.8cm}
\end{figure*}

\subsection{Ablation Study}
On both link prediction and node clustering tasks with Cora dataset, we validate the effectiveness of \underline{b}idirectional \underline{a}dversarial \underline{l}earning \textbf{(BAL)} and structure-aware \underline{p}rior \underline{d}istribution \underline{e}stimation \textbf{(PDE)}, respectively.
For such an ablation study, the basic setup about each subnet refers to the experiments above.
As shown in Table \ref{AS}, both BAL and PDE are equally important for our DBGAN to learn latent node representations. 
Specifically, compared with the baseline method `w/o both' without bidirectional adversarial learning and prior distribution estimation, the `w/o PDE' and `w/o BAL' receive obvious benefits for link prediction (e.g., 2.1\% and 1.5\% on AUC). Similarly, our DBGAN with both BAL and PDE achieves consistently the best performance over the three ablated methods. It can also be observed that there exists a performance decrease for `w/o BAL' on clustering task over `w/o both'. But the improvement of DBGAN implies that employing PDE facilitates BAL to great extent.

\begin{table}[t]
	\centering
	
	\caption{Effectiveness evaluation of BAL and PDE.}
	\begin{tabular}{@{}c|cc|ccc@{}}
		\toprule
		\multirow{2}{*}{\textbf{Methods}} & \multicolumn{2}{c|}{\textbf{Link Prediction}} & \multicolumn{3}{c}{\textbf{Clustering}} \\ \cmidrule(l){2-6} 
		& \textbf{AUC}               & \textbf{AP}               & \textbf{ACC}      & \textbf{NMI}      & \textbf{ARI}      \\ \midrule
		w/o both                     & 91.0                & 92.0               & 0.596       & 0.429       & 0.347       \\
		w/o PDE       & 93.1                & 93.9               & 0.684       & 0.472       & 0.431      \\
		w/o BAL     & 92.5                & 93.2               & 0.535       & 0.389       & 0.313       \\ \midrule
		DBGAN                    & 94.5                & 95.1               & 0.759       & 0.551       & 0.525       \\ \bottomrule
	\end{tabular}
	\label{AS}
	\vspace{-0.6cm}
\end{table}

\subsection{Efficiency Analysis}
\begin{figure}[t]
	\centering
	
	\subfigure[AUC]{
		\centering
		\includegraphics[width=1.6in]{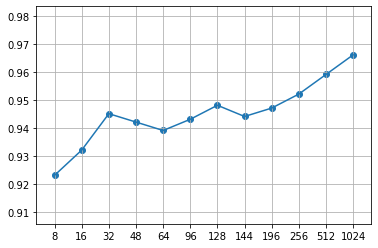}
	}%
	\subfigure[AP]{
		\centering
		\includegraphics[width=1.6in]{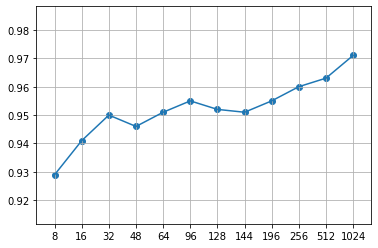}
	}%
	\caption{Impact of the dimension $q$ of learned latent representation on AUC and AP for link prediction task.}.
	\label{EA}
	\vspace{-0.8cm}
\end{figure}

The dimension of latent representation has a great effect on graph representation learning. To verify this fact, we vary the dimension of encoder output layer from 8-neuron to 1024-neuron for Cora dataset on link prediction task. 
The score achieved by DBGAN is shown in Figure\ref{EA}. 
Obviously, the performance of DBGAN keeps improving with dimension increasing. 
For a fair comparison, a low dimension is fixed in all our experiments. 
In particular, all our dimension is no more than 128, while the compared methods are generally opposite.
Even though in this case, we still achieve more promising results as in Tables~\ref{link prediction} and \ref{node cluster}.
This further verifies the dimension-efficient property of DBGAN. 
\subsection{Graph Visualization}
A promising unsupervised graph representation algorithm can usually preserve the original graph structure well in a low-dimensional space. 
To illustrate such a representation ability more intuitively, we use t-SNE\cite{tsne} to visualize the learned latent representations and original node features in a two-dimensional space.
Figure\ref{fig2} and Figure\ref{fig3} show the visualization results on Cora and Citeseer datasets respectively.
It can be seen that although our DBGAN performs graph representation learning in an unsupervised manner, it still can generate node representations that well preserve original adjacency relationships. 
Meanwhile, compared with raw features and the representations learned by DGI~\cite{DGI}, the results by our DBGAN is more discriminative, with smaller within-class scatter and larger inter-class scatter.
Specially, we can find that on Cora dataset, there exists many overlaps between pink and blue dots for GAE~\cite{GAE}, while such a phenomenon is alleviated greatly for our DBGAN.
\vspace{-0.2cm}
\section{Conclusion}
\vspace{-0.1cm}
In this paper, we propose a distribution-induced bidirectional adversarial learning network (named \textbf{DBGAN}) for graph representation learning. 
It is able to estimate the structure-aware prior distribution of latent representation via the learned prototypes, instead of the widely used Gaussian assumption, thus generating robust and discriminative representation of nodes.
More importantly, the generalization ability of our DBGAN is improved greatly while preserving representation ability, by balancing multi-level consistency with a bidirectional adversarial learning framework.
We have carried out extensive experiments on three tasks, and the results demonstrate the obvious superiority of our DBGAN over currently available alternatives in graph representation learning.
Our ongoing research work will extend our DBGAN to graph representation learning in the semi-supervised scenario.
\vspace{-0.1cm}
\section*{Acknowledgement}
This work was supported in part by Science and Technology Innovation 2030 - "New Generation Artificial Intelligence" Major Project under
Grant 2018AAA0102101, in part by the National Natural Science Foundation of China under Grant 61976018 and Grant 61532005, and in part by the Fundamental Research Funds for the Central Universities under Grant 2018JBZ001 and Grant 2019YJS048.

{\small
	\bibliographystyle{ieee_fullname}
	
}
\newpage

\end{document}